\documentclass[letterpaper, 10 pt, conference]{ieeeconf}  
\IEEEoverridecommandlockouts                              

\usepackage{microtype}
\usepackage{comment}

\usepackage{graphicx} 
\graphicspath{{./figures/}}
\usepackage{subcaption}
\usepackage{float}

\usepackage{amsmath} 
\usepackage{amssymb}  
\usepackage{units}
\usepackage{bm} 
\usepackage{multirow}

\usepackage{algorithm}
\usepackage{algpseudocode}

\makeatletter
\let\NAT@parse\undefined
\makeatother

\usepackage[square, numbers, sort&compress]{natbib}
\usepackage[hidelinks]{hyperref}
\usepackage{cleveref}

\usepackage{graphicx} 
\graphicspath{{./figures/}}
\usepackage{subcaption}
\usepackage[font=small]{caption}
\usepackage{xparse}
\usepackage{xcolor}
\usepackage{tabularx}
\usepackage{diagbox}
\usepackage{makecell}
\usepackage{soul}
\usepackage{natbib}

\title{\LARGE \bf Learning Vision-Based Bipedal Locomotion for Challenging Terrain} 

\author{Helei Duan, Bikram Pandit, Mohitvishnu S. Gadde, Bart van Marum, Jeremy Dao, Chanho Kim, Alan Fern
\thanks{*This work is supported by the NSF Award 2321851, NSF Award IIS-1849343 and DARPA Grant N66001-19-2-4035.}
\thanks{All authors are with Collaborative Robotics and Intelligent Systems Institute, Oregon State University, Corvallis, Oregon, 97331, USA. Email: \{{\tt\footnotesize duanh, panditb, gaddem, vanmarub, daoje, kimchanh, alan.fern}\}@oregonstate.edu.}
}

\begin{document}

\maketitle
\thispagestyle{empty}
\pagestyle{empty}

\begin{abstract}
Reinforcement learning (RL) for bipedal locomotion has recently demonstrated robust gaits over moderate terrains using only proprioceptive sensing. However, such blind controllers will fail in environments where robots must anticipate and adapt to local terrain, which requires visual perception. In this paper, we propose a fully-learned system that allows bipedal robots to react to local terrain while maintaining commanded travel speed and direction. Our approach first trains a controller in simulation using a heightmap expressed in the robot's local frame. Next, data is collected in simulation to train a heightmap predictor, whose input is the history of depth images and robot states. We demonstrate that with appropriate domain randomization, this approach allows for successful sim-to-real transfer with no explicit pose estimation and no fine-tuning using real-world data. To the best of our knowledge, this is the first example of sim-to-real learning for vision-based bipedal locomotion over challenging terrains.

\end{abstract}

\section{Introduction}
A robot's utility for useful work often hinges on its capacity to maneuver effectively across a spectrum of natural and structured terrains. For this purpose, bipedal robots have the potential to match human locomotion capabilities, but currently are far inferior. Approaching human performance requires a robot to perceive its surroundings, assess its states relative to the upcoming terrain, and dynamically adapt its gait. 
Robustly achieving such an integration of vision and locomotion remains an open problem for bipedal robots. 

Modern control approaches for vision-based legged locomotion \cite{Grandia2023PerceptiveControl, Fankhauser2018, Agrawal2021, Deits2015, Griffin2019FootstepTerrain, Mastalli2020, Jenelten2020, Gangapurwala2020} often decompose the problem into a control hierarchy, requiring robust whole-body control, footstep planning, accurate odometry estimation, and terrain mapping. Each level requires a set of modeling assumptions in order to process high-dimensional and raw proprioceptive and visual exteroceptive information. On the other hand, recent learning-based approaches make fewer modeling assumptions, often learning a direct mapping from high-dimensional sensory inputs to actuator commands. These approaches have shown strong empirical demonstrations of blind bipedal locomotion \cite{Siekmann2020b} and vision-based quadrupedal locomotion \cite{Yu2021, Miki2022LearningWild, Margolis2021, Agarwal2022LeggedVision, Rudin2022AdvancedEnd-to-End, Hoeller2023ANYmalRobots, Loquercio2022LearningSupervision} in real-world environments. However, they are typically trained in simulation and their success relies on designing techniques for reliable sim-to-real transfer, which can be particularly challenging when vision is one of the input modalities.

\begin{figure}
    \centering
    \includegraphics[width=0.97\linewidth]{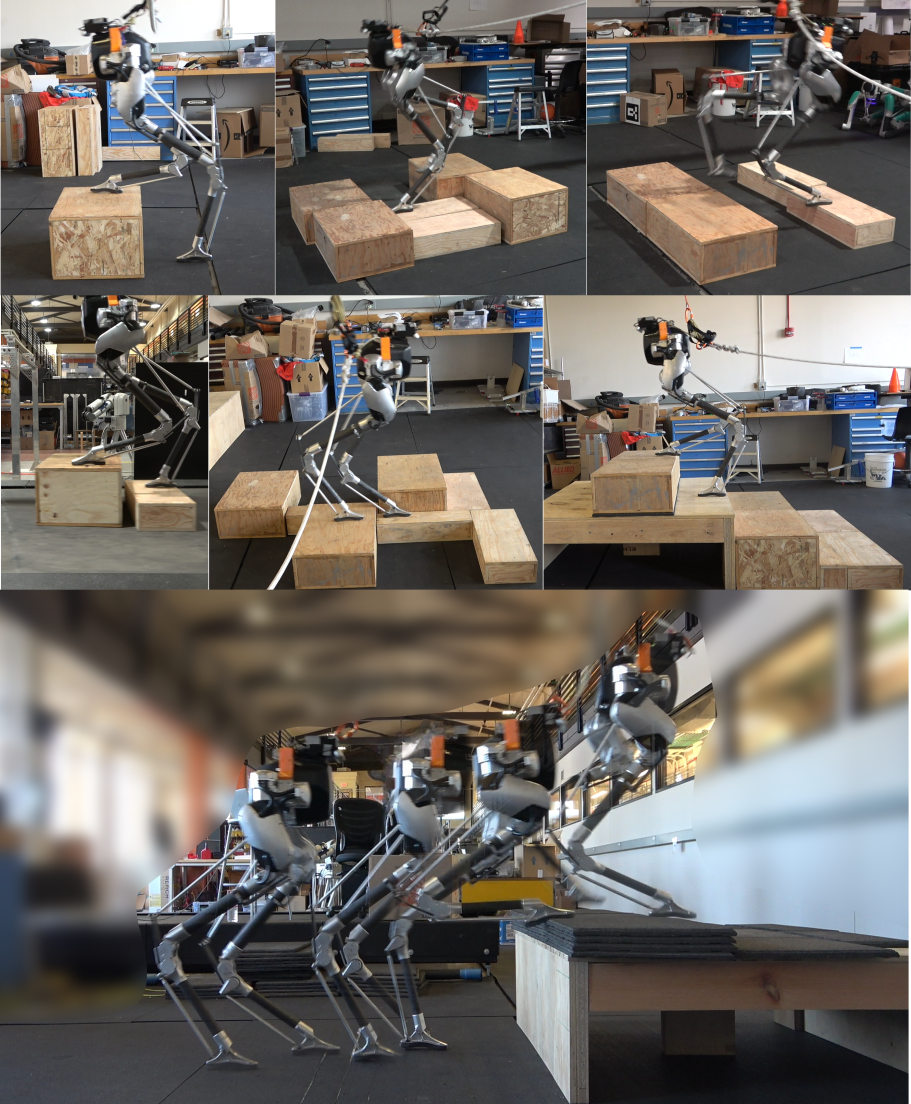}
    \caption{Our fully learned controller integrates vision and locomotion for reactive and agile gaits over terrains. The proposed approach enables bipedal robot Cassie traversing over challenging terrains, including random high blocks, stairs, 0.5m step up ($\sim$60\% leg length), with speed up to 1m/s.}
    \label{fig:lead}
\end{figure}

In this paper, we design and demonstrate a sim-to-real learning approach for vision-based bipedal locomotion over challenging terrain. To the best of our knowledge, this is the first such successful demonstration. A distinctive aspect of our approach is that it avoids the estimation of global odometry, which can be particularly challenging for legged locomotion due to estimation drifts from frequent contacts and aggressive motions. Instead, our approach learns to directly combine proprioceptive and visual information in the robot's local frame to make control decisions. In particular, our architecture is composed of two primary learned components: 1) a control policy whose input is proprioceptive information and a heightmap of a local region in front of the robot (Section \ref{sec:controller}), and 2) a heightmap predictor, which uses proprioceptive and egocentric depth images to predict a heightmap for the control policy (Section \ref{sec:terrain}). The key contribution of our work is the sim-to-real pipeline and the system integration for these components, which allows the overall locomotion controller to transfer successfully to the real world. In particular, we demonstrate the learned controller on a camera-equipped bipedal Cassie robot, which can traverse challenging terrains constructed in a lab environment shown in Figure \ref{fig:lead}. 

\section{Related Work} 

\textbf{Sim-to-Real Reinforcement Learning (RL) for Bipedal Robots.} Recently, RL-based controllers for bipedal locomotion have mostly focused on blind locomotion, where proprioception is the primary control input \cite{Siekmann2021, Duan2022LearningStones, Castillo2021, Li2023RobustLearning}. These works have produced controllers that can handle a limited amount of non-flat terrains. The learned policy \cite{Siekmann2021} tends to have a high cost of transport due to aggressive gaits and high swing trajectories that are required to maintain balance without knowledge of the local terrain. When traversing over challenging terrains, gait adaptation is highly required, and visual information becomes critical at certain times during the gait \cite{Matthis2013, GIBSON1958VisuallyAnimals}. The research question then is how to incorporate vision into RL policies so that bipedal robots can react to terrains and adapt their own gaits at the same time.  

\textbf{Learning Vision-based Legged Locomotion.} Previous work for quadrupedal robots has used RL methods to demonstrate successful sim-to-real transfer of vision-based locomotion controllers. One type of approach uses an elevation map or height scanners in the global reference frame as part of the RL policy input. For example, scattered elevation scans around each foot are used in quadrupeds \cite{Miki2022LearningWild} and bipeds \cite{vanMarum2023LearningTerrain}, while other methods \cite{Agarwal2022LeggedVision, Rudin2022AdvancedEnd-to-End, Hoeller2023ANYmalRobots, Rudin2022LearningLearning} use a uniformly structured elevation map around the robot, both of which require multiple sensors and careful calibration. In contrast, in this work we are interested in a solution that does not require careful calibration for global odometry. A second type of approach is to directly use vision inputs from a camera, such as depth images \cite{Agarwal2022LeggedVision, Fu2021, Zhuang2023RobotLearning, Yu2021} or RGB images \cite{Haarnoja2023LearningLearning}, as the inputs to a RL policy. This end-to-end training is often carried out via teacher-student training \cite{Agarwal2022LeggedVision, Zhuang2023RobotLearning}, which can exploit the teacher's access to privileged information in simulation. While these approaches have been successful on quadrupeds on hardware, it is unclear how well they can work for bipeds in the real world, where the contact locations become critical for stability and unintended contact forces with the ground can much more easily tip over the robot.

\textbf{Local Terrain Mapping for Legged Robots.} 
Model-based terrain mapping techniques have shown successful deployment onto hardware via odometry and fusion of multiple visual sensors \cite{Fankhauser2016a, Fankhauser2018ProbabilisticLocalization, Plagemann2009ALocomotion}. These techniques strongly rely on pose estimation of the floating base in the global frame, where the map can drift due to inaccuracies from pose estimation. Previous visual-based quadrupedal locomotion work \cite{Miki2022LearningWild} reported that large amounts of domain randomization are required to overcome the noises and drifts from such mapping techniques. On the other hand, recent learning-based techniques \cite{Hoeller2022NeuralTerrain, Yang2023Real-TimeEstimations} have shown promising results when reconstructing the terrains from multiple cameras, but the use of robot global pose is still required. In this paper, our focus is on responding to terrain changes in front of the robot, for which we use a single depth camera to provide an egocentric view of the terrain. Along with robot states, the reconstructed heightmap is entirely in the robot's local frame. Our method removes the need to use global estimation when the robot has to react to local terrains rapidly.

\section{System Overview}
\label{sec:overview}
\begin{figure}[h!t]
\centering
\includegraphics[width=0.99\columnwidth,angle=0]{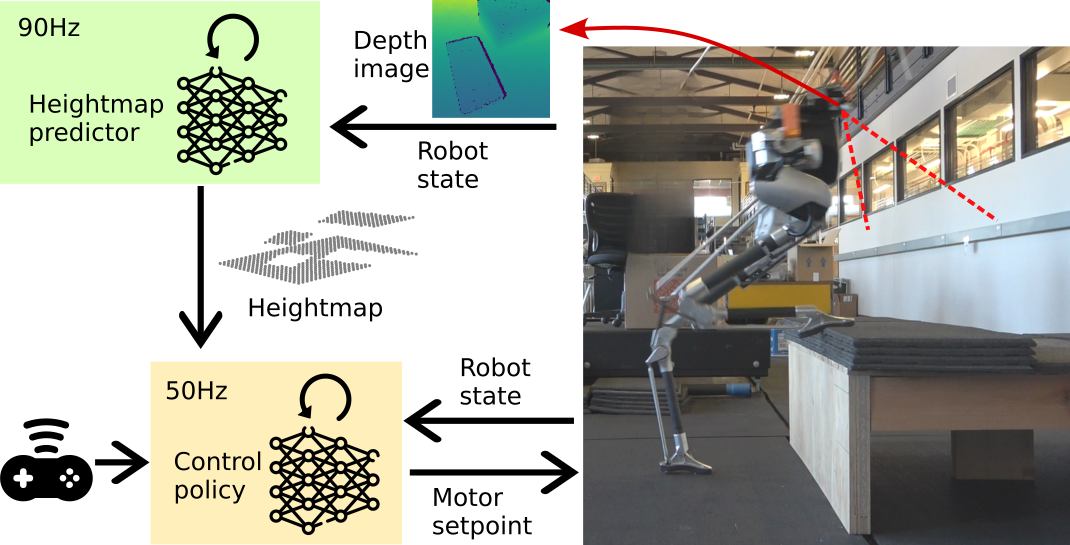}
\caption{Overview of the locomotion policy with vision module.}
\label{fig:overview}
\end{figure}

Figure \ref{fig:overview} illustrates our overall system, which has two main components: 1) a locomotion policy, which outputs PD setpoints for the robot actuators based on proprioception, a local terrain heightmap, and user commands, and 2) a heightmap predictor, which outputs a predicted heightmap based on proprioceptive information and images from a depth camera. These components are learned in simulation and then transferred to the real robot.

The training pipeline first uses RL to train the locomotion policy in simulation. This training process randomizes the terrain, user commands, and physical parameters to help with sim-to-real transfer. Next, using the learned policy, the pipeline collects data from a simulated depth camera attached to the robot, which is paired with ground-truth heightmap information to be used for supervised learning of the heightmap predictor. Training this predictor also involves domain randomization and added noise to facilitate sim-to-real transfer. Sections \ref{sec:controller} and \ref{sec:terrain} describe the architectural and training details of each component. 


\section{Learning a Terrain-Aware Locomotion Policy}
\label{sec:controller}

The main control objective is to follow speed and heading commands while maintaining balance over possibly challenging terrains. 
Below, we describe the observation space, action space, architecture of the policy, and training methods. 


\subsection{Control Policy Design}

\noindent \textbf{Observation Space.} The policy input includes: 1) \emph{proprioceptive information} containing the orientation (in quaternion) and angular velocity of the floating base, and position and velocity for all measurable actuated and unactuated joints, 2) 
\emph{terrain heightmap} from a 1.5m by 1m area in front of the robot at a 5cm resolution (see Figure \ref{fig:terrain}), which encodes the ground height at each point relative to the robot's floating base. The relative encoding means that the heights vary as the robot moves up and down during its gait, but enables us to avoid using global mapping and odometry estimation techniques, 3) \emph{user commands}, which include X and Y linear velocities along with direction and rotational velocity around the robot's yaw axis, and 4) \emph{periodic clock}, as used in prior work on locomotion control \cite{Siekmann2020b}, which consists of two functions for each leg, $\sin{(2\pi(\phi_t+\gamma^{i}_t))}$ and $\cos{(2\pi(\phi_t+\gamma^{i}_t))}$. Here $i\in[\textit{left, right}]$ indicates the leg, $\phi$ is a monotonically increasing phase variable that is incremented as $\phi_{t+1}=\phi_{t}+\Delta\phi_t$, so that $\Delta\phi_t$ varies the gait frequency, and [$\gamma^{left}_t, \gamma^{right}_t$] are period shifts that alter the clock values for each leg, thus changing the contact sequence.

\noindent \textbf{Action Space.} The RL policy operates at 50Hz and outputs PD setpoints for all motors, which are provided to a PD controller operating at 2kHz. To enable gaits that can more flexibly adapt to the terrain, the RL policy also outputs three extra values, representing the clock increment $\Delta\phi_t$ and residuals of shifts for both legs [$\Delta\gamma^{left}_t, \Delta\gamma^{right}_t$]. 

\begin{figure}[t]
\vspace*{0.3cm}
\centering
\includegraphics[width=0.85\columnwidth,angle=0]{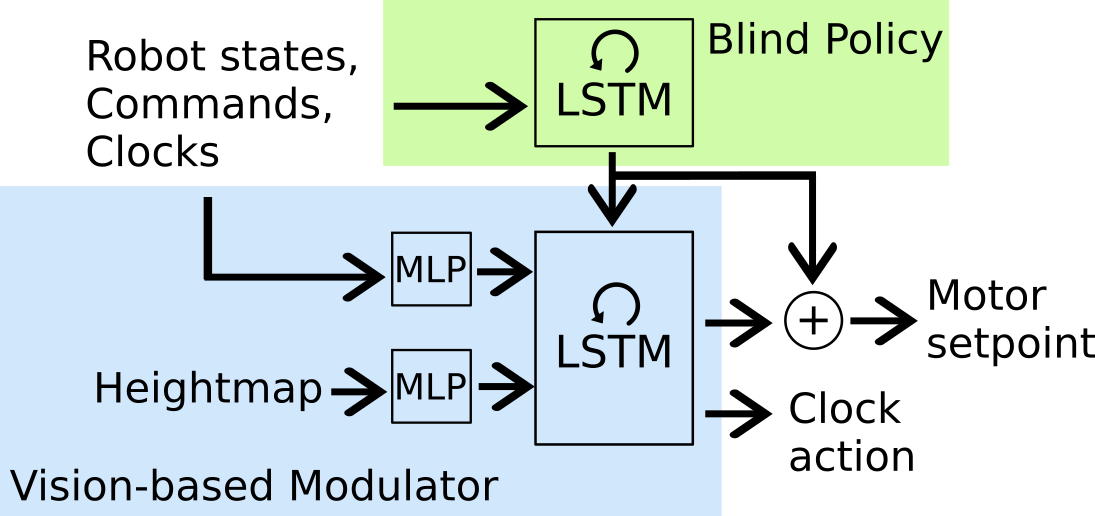}
\caption{Policy consists of a blind policy and a vision-based modulator. }
\label{fig:policy}
\end{figure}
\noindent \textbf{Policy Architecture.} We use a neural network to represent the policy for mapping observation sequences to actions. The policy architecture (Figure \ref{fig:policy}) contains two main components, a pretrained \emph{blind policy} and a \emph{vision-based modulator}. The architecture motivation is for the blind policy to provide a baseline locomotion control signal, which is effective for moderate terrains. For more complex terrain, the vision-based modulator is then able to adjust the baseline control based on details of the local terrain. 

The \emph{blind policy} is based on prior work \cite{Siekmann2020b} and uses an LSTM network to produce actions. This policy uses all of the available inputs, except for the heightmap information, and is trained on relatively flat ground with various gait frequencies and shifts. The resulting policy is robust to moderate disturbances. The input to the \emph{vision-based modulator} includes all of the available observations, including the heightmap, in addition to the action produced by the blind policy. This modulator outputs a ``modulating action" as well as clock actions to modify the clock parameters.

\subsection{Policy Training}

The policy is trained via the PPO actor-critic RL algorithm \cite{Schulman2017} in a MuJoCo simulation environment \cite{Todorov2012} where the robot aims to follow randomized commands over randomized terrains. The training is conducted using 80 cores on a dual Intel Xeon Platinum 8280 server on the Intel vLab Cluster. We used two modifications to standard PPO that helped speed up and stabilize learning. First, we modified the PPO loss function to include a mirror loss over robot propriocepive inputs as well as visual inputs. This loss encourages the policy to choose symmetric actions when facing the same terrain that is symmetrically mirrored about the saggital plane. Second, since our reward function (described below) uses privileged information, we found it useful to provide the critic with privileged inputs on addition to the observations input to the policy. These privileged inputs include the robot height and feet positions in the global frame, a square 0.5m heightmap around each foot, and two rangefinder sensors from the tip of each foot. This privileged information provides the critic with a more accurate 3D picture around each foot, which helps it more accurately predict future rewards, such as unfavorable collision events with the terrain, as shown effective in \cite{Ma2023LearningManipulators}.

\begin{figure}
\vspace{0.1cm}
\centering
\includegraphics[width=0.95\columnwidth,angle=0]{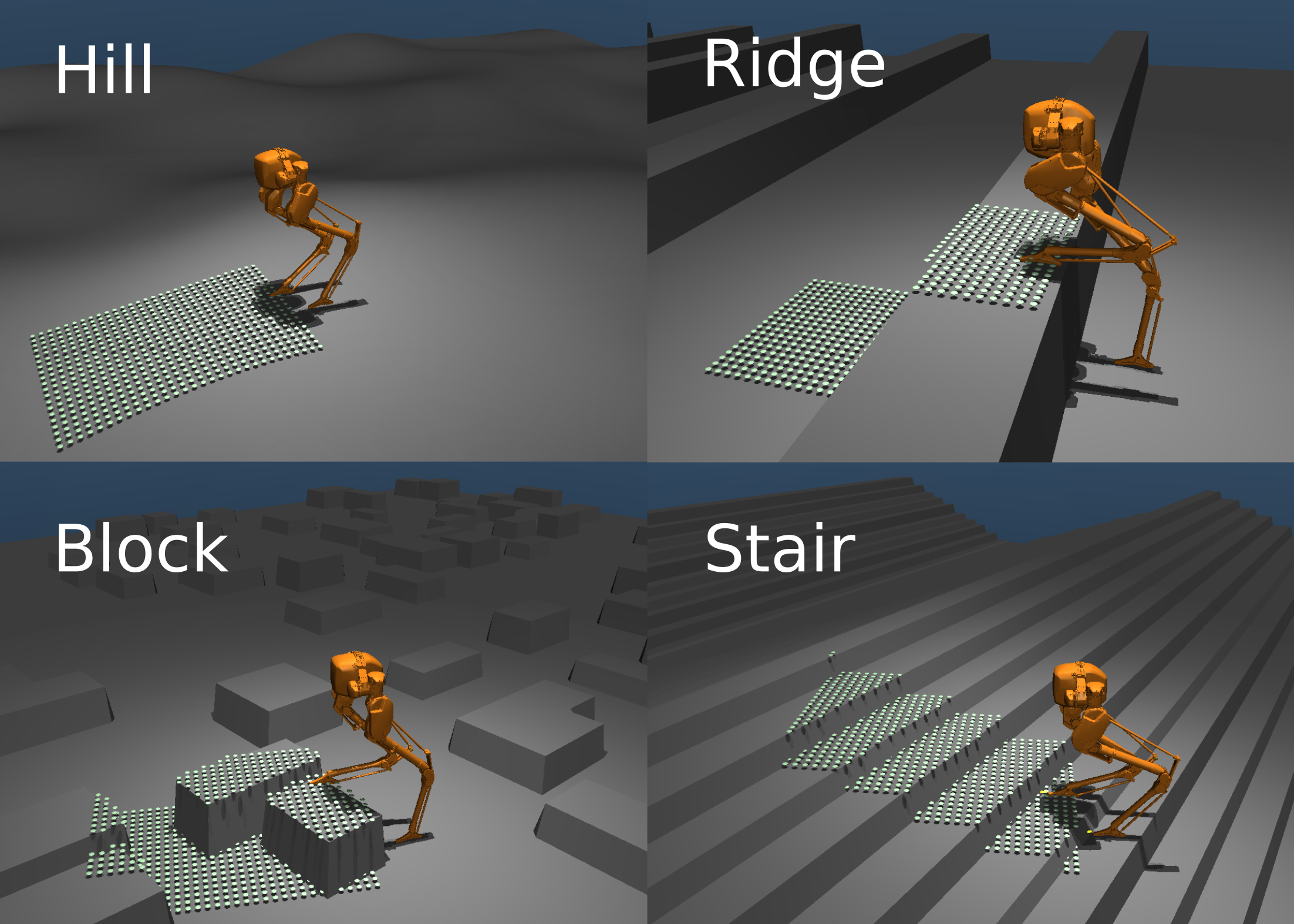}
\caption{Types of terrain used in training.}
\label{fig:terrain}
\end{figure}

\noindent\textbf{Training Episode Generation.} Each training episode involves a randomly generated terrain map of size 20m x 20m. Each map is one of 5 terrain types, illustrated in Figure \ref{fig:terrain}, that are reflective of different human-centric environments. These include: 1) flat - the easiest terrain with no features, 2) hills - a smoothly varying heightmap, 3) blocks - randomly located blocks of random lengths, widths, and heights, 4) ridges - a sequence of random width and height ridges that span the length of the map, and 5) stairs - upward and downward stairs of varying width and height. The randomization parameters of each terrain type are listed in Table \ref{fig:terrain}, and terrains are generated in a way to avoid intersecting features. 
Each episode selects a single terrain type according to the following probability distribution $[0.03, 0.07, 0.35, 0.2, 0.35]$, which puts the majority of the probability mass on the three most difficult terrains. This allows the policy to gain some experience on easier terrains, which is useful early in learning, but focuses most of the learning effort on more difficult terrain that requires careful integration of vision and locomotion. We found the key to train a robustness and non-aggressive control policy relies on the terrain generation distribution rather than iterating and adding heuristic-based reward terms. For example, stairs naturally regulate the step length and ridges regulate step height via the heightmap representation. 

Given a terrain map, each training episode starts with the robot being randomly spawned in a standing position near the center of the map and facing a random direction. Next a random command is given to the policy from a list including: step-in-place, step-in-place-turn, walk, walk-turn, with a sampling probability of [0.05, 0.05, 0.6, 0.3]. The commanded X, Y, and turn velocities are uniformly sampled from the ranges [-0.5, 1.0]m/s, [-0.3, 0.3]m/s, and [-22.5, 22.5] degrees/s,  respectively. When the robot is on the ridge, stair, or block terrain types, the velocity commands exclude backward and sideways movement, due to using only a single forward facing camera. After the initial command, once during each episode the command is randomly changed at a time randomly sampled from [200, 250] timesteps. 

Each episode runs for a maximum of 400 timesteps, which is 8 seconds of simulated time. Episode termination conditions include: 1) roll or pitch angle of the floating base is greater than 15 degrees; 2) the norm of linear velocities of the base is greater than 1 plus the commanded velocity; 3) the robot base height is below 40cm from robot base to terrain; 4) the robot's body collides with the terrain. The conditions correspond to undesirable robot behavior and implicitly punish the robot by causing it to not receive future rewards.

\begin{table}[]
\vspace*{0.1cm}
\centering
\caption{Ranges for terrain randomization used in training and evaluation. All terrains are uniformly sampled within the range during training or evaluation.}
\label{tab:terrain-rand}
\begin{tabular}{c|c|c||cc}
\hline
\multirow{2}{*}{\textbf{Terrain}} & \multirow{2}{*}{\textbf{Parameter}}                        & \multirow{2}{*}{\textbf{Range}} & \multicolumn{2}{c}{\textbf{Range for Evaluation}}     \\ \cline{4-5} 
                                  &                                                            &                                 & \multicolumn{1}{c|}{\textit{Easy}}   & \textit{Hard}  \\ \hline
Ridge                              & height {[}m{]}                                             & {[}0.05, 0.6{]}                 & \multicolumn{1}{c|}{{[}0.05, 0.5{]}} & {[}0.5, 0.6{]} \\ \hline
\multirow{3}{*}{Stair}            & height {[}m{]}                                             & {[}0.05, 0.2{]}                 & \multicolumn{1}{c|}{{[}0.05, 0.1{]}} & {[}0.1, 0.2{]} \\ \cline{2-5} 
                                  & length {[}m{]}                                             & {[}0.25, 0.4{]}                 & \multicolumn{1}{c|}{{[}0.4, 0.4{]}} & {[}0.25, 0.4{]} \\ \cline{2-5} 
                                  & steps & {[}4, 28{]}                     & \multicolumn{1}{c|}{{[}4, 12{]}}     & {[}12, 28{]}   \\ \hline
Block                          & \begin{tabular}[c]{@{}c@{}}length or \\ width\end{tabular} & {[}0.4, 1{]}                    & \multicolumn{1}{c|}{{[}1, 1{]}}      & {[}0.4, 1{]}   \\ \cline{2-5} 
\textbf{}                         & height {[}m{]}                                             & {[}0.05, 0.4{]}                 & \multicolumn{1}{c|}{{[}0.05, 0.2{]}} & {[}0.2, 0.4{]} \\ \hline
\end{tabular}
\end{table}

\noindent\textbf{Reward Function.} Our reward function aims to produce a gait with a well regulated contact sequence that is able to traverse the simulated environment in a way that is likely to transfer to a real robot. In particular, we use a three component reward function where all components are weighted equally, $R = R_0 + R_{\text{accel}} + R_{\text{collision}}$. The base locomotion component $R_0$ is the reward function used in prior work on blind locomotion \cite{Siekmann2020b}, which regulates the contact sequence and timing. This reward component encourages alternating swing and stance phases to align with the clock values provided to the policy. 

Besides the base locomotion reward, we identified additional components as being important for facilitating sim-to-real transfer for complex terrains. The foot acceleration component $R_{\text{accel}}$ penalizes the left and right foot accelerations $\Ddot{x}_l$ and $\Ddot{x}_r$. This reward helps prevent fast swing leg motions, which we found could arise during training on difficult terrain. Specifically, the reward is defined as $R_{\text{accel}} = 0.05\exp{\left(-0.02 \cdot \left( \|\Ddot{x}_l\| + \| \Ddot{x}_r\|\right)\right)}$, which provides a more positive reward for smaller accelerations. The foot collision component $R_{\text{collision}}$ adds a negative penalty of -5 whenever the forefront of the foot is stumbled by the terrain, which helps to achieve collision-free leg-swing trajectories. However, due to the nature of RL, this term only acts as a soft constraint that the robot may violate in favor of not falling down, in order to collect more reward throughout the training episode. We found training without this penalty term can work in simulation, but results in significantly more frequent foot stumble and collision events, and subsequently prevents sim-to-real transfer. Touch sensors are added at the forefront of each foot's collision model to detect collision events and trigger the negative reward.

\noindent\textbf{Domain Randomization.} To enable successful sim-to-real transfer and diversify the data distribution during training, the training process involves a range of domain randomization, shown in Table \ref{tab:policy-dr}, over model parameters, actuation parameters, visual inputs, and delays. The model parameters are randomized per episode to simulate a range of robot models and also provide a wide range of state space that the policy can learn from. We found that randomizing the torque efficiency parameter is particularly important for sim-to-real on extreme terrain, such as a 0.5m step up, due to the torque saturation of the knee motor. In addition, the torque command sent to the simulator is delayed randomly up to 3ms. The visual inputs are randomized to simulate noise from the heightmap estimator and prevent the policy from over-fitting to the exact simulated heightmaps. Prior work \cite{Miki2022LearningWild} based on teacher-student training also found that heightmap randomization was important for sim-to-real transfer. The entire heightmap is shifted per episode and policy step  in all directions to simulate temporal noises. The heightmap is passed into the policy with a randomized amount of delay up to 100ms, in order to account for faster locomotion speeds.

\begin{table}[]
\vspace*{0.1cm}
\centering
\caption{Parameters and ranges used in domain randomization. All parameters are uniformly sampled within the range.}
\label{tab:policy-dr}
\resizebox{\columnwidth}{!}{%
\begin{tabular}{cl|l|l}
\hline
\multicolumn{2}{c|}{\textbf{Parameters}}                                                                                             & \textbf{Range}                                                                                                     & \textbf{Unit}   \\ \hline
\multicolumn{1}{c|}{\multirow{6}{*}{\begin{tabular}[c]{@{}c@{}}Simulation\\ Model\end{tabular}}} & Joint Damping            & {[}0.5, 2.5{]}                                                                                            & \%     \\ \cline{2-4} 
\multicolumn{1}{c|}{}                                                                            & Mass                     & {[}-0.25, 0.25{]}                                                                                         & \%     \\ \cline{2-4} 
\multicolumn{1}{c|}{}                                                                            & Center of Mass Location          & {[}-0.01, 0.01{]}                                                                                         & m      \\ \cline{2-4} 
\multicolumn{1}{c|}{}                                                                            & Passive Spring Stiffness & {[}-500, 500{]}                                                                                           & Nm/rad \\ \cline{2-4} 
\multicolumn{1}{c|}{}                                                                            & Torque Efficiency        & {[}0.9, 1.0{]}                                                                                            & \%     \\ \cline{2-4} 
\multicolumn{1}{c|}{}                                                                            & Torque Delay             & {[}0.5, 3{]}                                                                                       & ms      \\ \hline
\multicolumn{1}{c|}{\multirow{3}{*}{Heightmap}}                                                  & Shift in XY              & \begin{tabular}[c]{@{}l@{}}{[}-0.05, 0.05{]} per episode\\ {[}-0.05, 0.05{]} per policy step\end{tabular} & m      \\ \cline{2-4} 
\multicolumn{1}{c|}{}                                                                            & Shift in Z               & \begin{tabular}[c]{@{}l@{}}{[}-0.1, 0.1{]} per episode\\ {[}-0.02, 0.02{]} per policy step\end{tabular}   & m      \\ \cline{2-4} 
\multicolumn{1}{c|}{}                                                                            & Delay                    & {[}20, 100{]}                                                                                             & ms     \\ \hline
\end{tabular}%
}
\end{table}

 \section{Heightmap Prediction from Egocentric Vision}
\label{sec:terrain}
The heightmap predictor is a neural network that maps a history of egocentric depth images and robot states into an estimated heightmap in front of the robot. This problem is challenging due to the aggressive camera motions, occlusions, and noisy images. Below, we describe the architecture and simulation-based training process used to achieve successful sim-to-real transfer.

\begin{figure}[h!tbp]
\vspace{-.2cm}
\centering
\includegraphics[width=0.9\linewidth,angle=0]{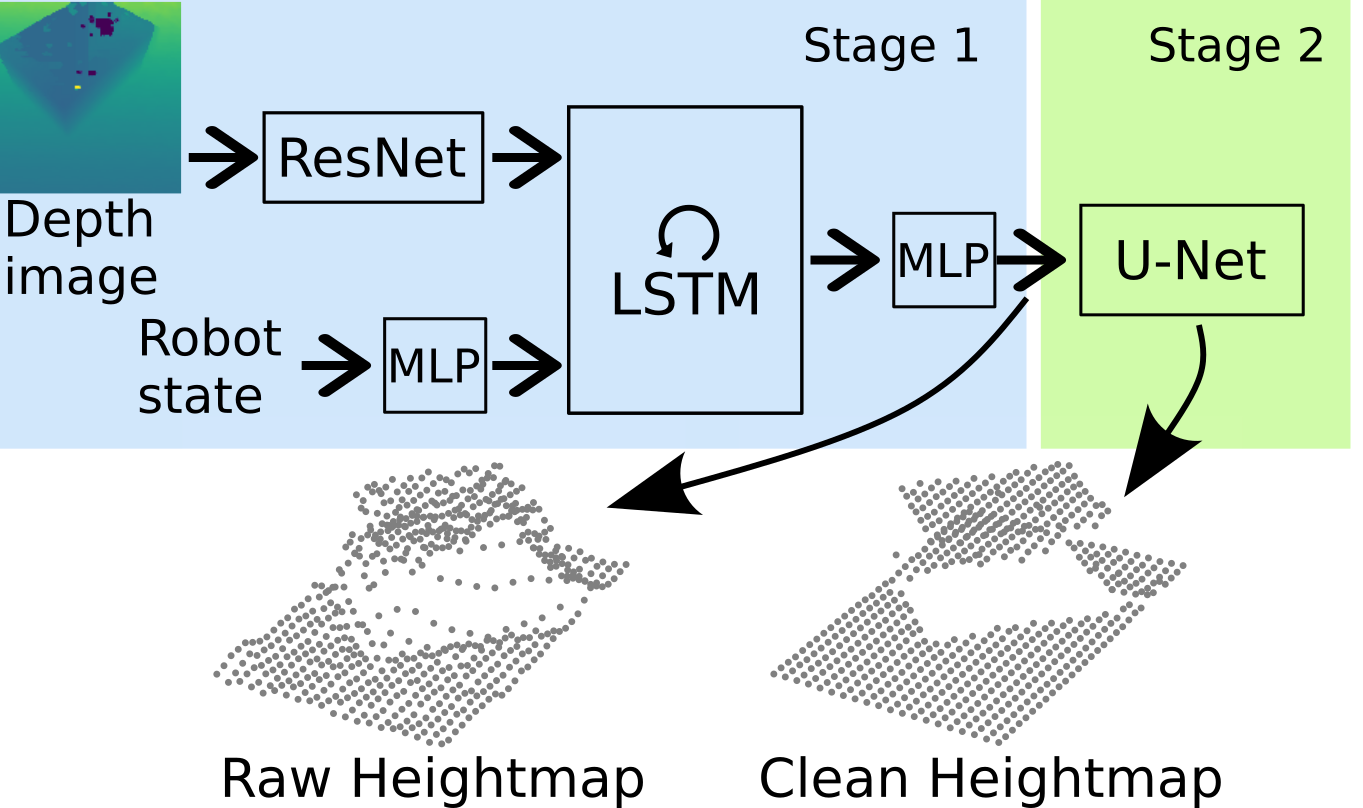}
\caption{Predictor architecture. Heightmap is captured from hardware.}
\label{fig:predictor}
\vspace{-.55cm}
\end{figure}

\noindent\textbf{Network Architecture and Losses}
Figure~\ref{fig:predictor} shows the network architecture, which consists of two stages. For the first stage, we use an LSTM network so that the memory can help reconstruct missing information from the history of robot states and depth images.
Training of this stage is done by minimizing the mean-squared error to the ground truth heightmap with heights relative to the robot floating base. We found that training just Stage 1 resulted in heightmaps with non-flat surfaces around corners and edges, which were difficult to correct via straightforward architecture and loss function modifications. To improve over the raw heightmaps, Stage 2 utilizes a U-Net architecture \cite{Ronneberger2015U-net:Segmentation} which has shown to be effective in learning a pixel-to-pixel transformation. This stage takes in the raw heightmap and outputs a cleaned version of the same size. Stage 2 uses L1 loss and leads to a refined heightmap with sharper corners, edges, and flat surfaces. 

\noindent\textbf{Simulation-Based Data Generation}
Given a trained locomotion control policy, we use the simulator to execute episodes of the policy to collect training data. In particular we run the policy in stochastic mode, where the actions are sampled from the action distribution at each time step. This has the effect of producing data around the typical data distribution the policy will encounter at runtime. The dataset contains one trajectory for each policy-execution episode, where the trajectory at time $t$ stores the robot state $s_t$, the depth image $I_t$ and the ground truth heightmap $m_t$. The range of domain randomization is the same as it is during training. Simulation model adds an egocentric camera to generate depth images at each step. The pose is at the top of the floating-base and tilted by 60 degrees from horizontal plane, giving the range of view of approximately 2.5m$\times$2.5m in front of the robot, which is sufficient for the size of heightmap. The depth images are rendered in MuJoCo with size of 848$\times$480, and then resized into 128$\times$128. Robot states include feet positions in robot's local frame. The final training set contains 30,000 episodes. 


\noindent\textbf{Depth Image Randomization}
Depth images from real cameras are significantly more noisy than the clean depth images from simulated cameras. During the generation of training episodes we add a set of randomizations to the generated depth images to help bridge the sim-to-real gap. First, the camera pose and field of view (FOV) are randomized per episode. Camera pose shift has a range of $\pm$1 cm for XYZ and $\pm$1 degree for pitch angle. FOV shift has a range of $\pm$1 degree. After data collection, a post-processing step adds a set of random noises on top of the rendered depth images, including Gaussian noise, image rotation, edge noise, random small objects on the image, and spot noise. Each type of noise is included in an image with a probability of 0.3. We found that this combination of depth image randomization allowed sim-to-real for the learned predictor more effectively. 


\begin{figure}[h!tbp]
\centering
\vspace*{0.3cm}
\includegraphics[width=0.72\linewidth,angle=0]{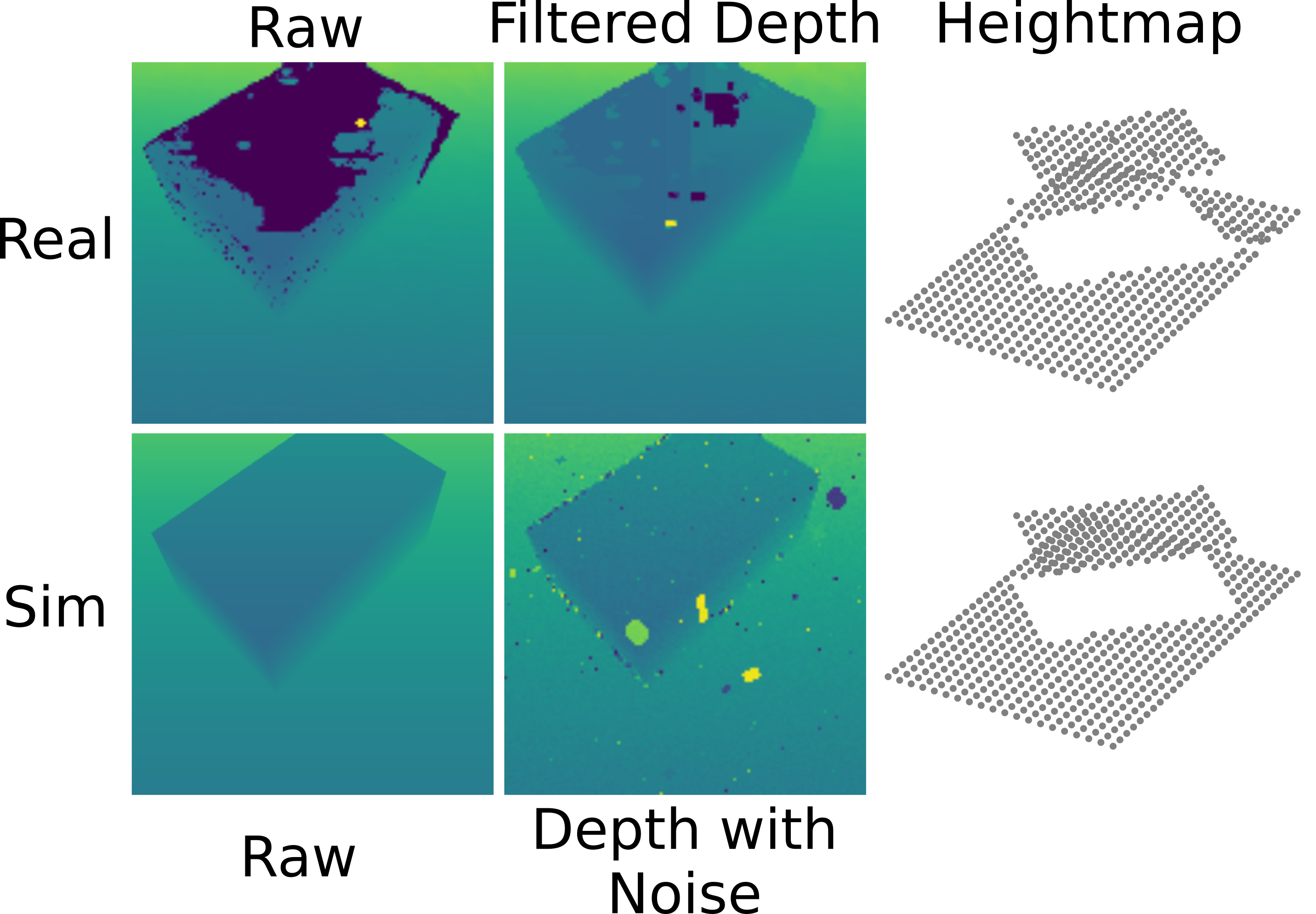}
\caption{Depth image from simulation and real world, with corresponding real predicted heightmap and simulation heightmap.}
\label{fig:depth}
\end{figure}

\section{Simulation Results}
\subsection{Policy Performance}
\begin{figure*}
\vspace{0.1cm}
\centering
\includegraphics[width=0.95\textwidth,angle=0]{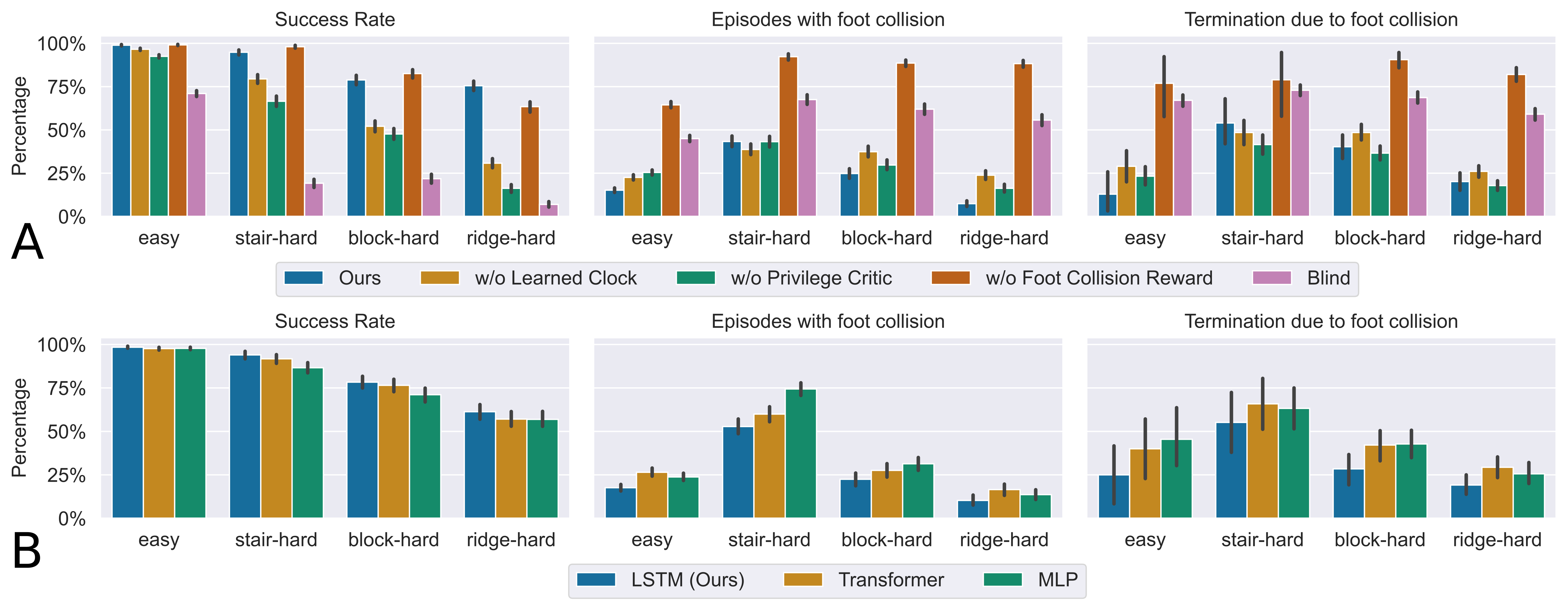}
\caption{\textbf{A}. Ablation study on policy with simulation heightmap. \textbf{B}. Ablation study on policy with different heightmap predictor architectures. Each ablation study uses data collected from a range of terrains defined in Table \ref{tab:terrain-rand}. \textbf{Success rate} indicates the robot does not fall down for 10 seconds of rollouts. \textbf{Episodes with foot collision} indicates the number of episodes that have one or more foot collision events occurred during rollouts, and such random collision events are unfavorable towards hardware deployment. \textbf{Termination due to foot collision} shows the percentage of foot collision events that lead to failures. All plots are evaluated with a confidence interval of 95\%.}
\label{fig:results}
\vspace{-0.7cm}
\end{figure*}

We use the trained policy, along with a number of different policy setups, to evaluate the performance in simulation for the ablation study. For each terrain we define an easy and a hard configuration as shown in Table \ref{tab:terrain-rand}. For each policy setup, we collect 1000 episodes per terrain mode and compute three metrics as shown in Figure \ref{fig:results}-A. \textit{Ours} is the policy trained using all setups and is for sim-to-real. Other setups are for controlled tests by removing one feature at a time. We also trained a blind policy across all terrains. In \textbf{Success Rate}, all policies have approximately the same performance at easy mode of terrains. For more difficult terrain modes, however, policies \textit{w/o Learned Clock} and \textit{w/o Privilege Critic} show significantly lower success rates. This means that control over the gait's contact timing and sequence as well as a more accurate value estimation improve policy performance over hard terrains. \textbf{Episodes with foot collision} shows that, compared to \textit{Ours}, other policies have significantly more foot collisions events. These random foot collisions with the terrain could lead to failures. Indeed, \textbf{Terminations due to foot collision} indicates that collisions account for most failure cases overall. Although foot collisions lead to frequent failures, policy \textit{w/o Foot Collision Reward} has a similar success rate as \textit{Ours}. When looking at policy \textit{w/o Foot Collision Reward} in simulation, the policy learns to deal with collisions and treat the collisions as potentially useful proprioceptive feedback when the robot touches terrains. For example, the robot will walk up high step-ups by sliding the foot along the vertical surface of the terrain. \textit{Blind} policy has the most aggressive swing leg motions to account for non-observable terrains and has the worst performance.

\subsection{Local Terrain Reconstruction}
We evaluated the heightmap predictor in simulation to validate the quality of the reconstruction shown in Table \ref{tab:predictor-recon}. We also implemented other architectures to use for ablations, including an MLP model and a transformer-based model, and they all have robot states and depth image as the input. A key distinction among the model architectures is the history representation. \textit{LSTM} has implicit history, \textit{Transformer} has a fixed window size of 0.6 seconds to allow reasonable inference during runtime, and \textit{MLP} does not have history. Additionally, the comparison also includes an LSTM model, \textit{LSTM w/o robot states}, that does not use robot states as input. Among all models, \textit{LSTM} achieves the best reconstruction loss, likely benefited from the implicit representation on history which allows useful extraction from robot states with egocentric view. Without robot states as input, \textit{LSTM w/o robot states} produces worse reconstructions, indicating the requirement of proprioception and vision together to accurately estimate local terrain. 
\begin{table}[ht]
\centering
\caption{Reconstruction loss with various heightmap predictors.}
\label{tab:predictor-recon}
\resizebox{\columnwidth}{!}{%
\begin{tabular}{|l|l|l|}
\hline
\multicolumn{1}{|l|}{Model Architecture} & Reconstruction Loss (MAE) [cm]  \\ \hline
LSTM                                  & \textbf{2.806} \\ \hline
Transformer                           & 4.221 \\ \hline
MLP                                   & 4.932 \\ \hline
LSTM (w/o robot states)               & 4.448 \\ \hline
\end{tabular}%
}
\end{table}

\subsection{Closed-loop Evaluation}
We use each variation of the heightmap predictor to couple with the policy (\textit{Ours} in Figure \ref{fig:results}-A) to evaluate the closed-loop performance in simulation shown in Figure \ref{fig:results}-B. We use the same metrics used in policy performance. During each rollout, the predictor takes in the noisy depth images from simulation rendering and the policy takes in the clean heightmap from the output of the Stage in Figure \ref{fig:predictor}. In \textbf{Success Rate}, all predictors produce similar performance over each terrain mode. This potentially means that, regardless of the errors from reconstruction, the learned policy is robust enough to estimation noises. In \textbf{Episodes with foot collision}, compared to \textit{LSTM}, other models show worse performance and produce more collision events. In \textbf{Termination due to foot collision}, compared to \textit{LSTM}, other models fails with higher chances from unfavorable foot collisions.

\section{Sim-to-real Transfer}
\subsection{Experimental Setup}
We deployed the proposed system on the bipedal robot Cassie. To endow the vision system with fast inference, we added a D455 Intel Realsense camera and an NVIDIA Jetson Orin Nano module on Cassie. Customized mount with camera is calibrated by matching depth images with simulated images under the same robot pose. Depth images are post-processed with a hole-filling filter and distance clipping before being sent into the heightmap predictor. The main control policy is running on the robot's main computer and inference of the heightmap predictor is running asynchronously on the Jetson. The camera depth stream is set to 90 FPS and heightmap prediction inference alone can run at 200Hz. The end-to-end delay between the main control policy and the heightmap predictor is measured up to 20ms, including UDP communication delay, camera stream, and model inference. To create structured terrains similar to those in simulation training, we built wooden blocks with various heights and sizes in the lab environment shown in Figure \ref{fig:lead}.

\subsection{Evaluation}
We tested the proposed system over various terrains. Overall, the system is able to control robot Cassie going over single high blocks, stairs, and random blocks, with various dimensions. Although the egocentric view of the camera cannot look underneath the robot, we found the policy is able to go over terrains while reasoning about the swing trajectory for the terrain underneath it. This observation potentially means that the policy keeps an internal odometry when approaching upcoming terrains. Also, the system enables the robot to go over a 0.5m high step-up, where the torque saturation happens on the leading stance leg when lifting the entire robot up from lower ground to a high step. We also tested the learned system on a treadmill while a human operator continually fed random blocks down the treadmill. The robot is able to go over the upcoming blocks and maintain travel speed and heading. Please refer to the submission video for hardware demonstrations.

\section{Conclusion}
In this work, we proposed a fully learned visual-locomotion system using neural networks. To deal with constrained locomotion over complex terrains, we used simulation to train a robust control policy, considering adaptive gaits and collision-free swing leg. A vision module is trained using simulation-only data to convert an egocentric view from a single depth camera to a terrain heightmap in robot local frame. The entire system ran onboard and achieved successful sim-to-real transfer without the need for explicit odometry estimation. We believe these key components are fundamental to our study as well as future research on learning vision-based dynamic locomotion for legged robots. 





\clearpage
\def\bibfont{\footnotesize}
\bibliographystyle{IEEEtranN}
\bibliography{references}

\begin{thebibliography}{36}
\providecommand{\natexlab}[1]{#1}
\providecommand{\url}[1]{#1}
\csname url@samestyle\endcsname
\providecommand{\newblock}{\relax}
\providecommand{\bibinfo}[2]{#2}
\providecommand{\BIBentrySTDinterwordspacing}{\spaceskip=0pt\relax}
\providecommand{\BIBentryALTinterwordstretchfactor}{4}
\providecommand{\BIBentryALTinterwordspacing}{\spaceskip=\fontdimen2\font plus
\BIBentryALTinterwordstretchfactor\fontdimen3\font minus
  \fontdimen4\font\relax}
\providecommand{\BIBforeignlanguage}[2]{{%
\expandafter\ifx\csname l@#1\endcsname\relax
\typeout{** WARNING: IEEEtranN.bst: No hyphenation pattern has been}%
\typeout{** loaded for the language `#1'. Using the pattern for}%
\typeout{** the default language instead.}%
\else
\language=\csname l@#1\endcsname
\fi
#2}}
\providecommand{\BIBdecl}{\relax}
\BIBdecl

\bibitem[Grandia et~al.(2023)Grandia, Jenelten, Yang, Farshidian, and
  Hutter]{Grandia2023PerceptiveControl}
\BIBentryALTinterwordspacing
R.~Grandia, F.~Jenelten, S.~Yang, F.~Farshidian, and M.~Hutter, ``{Perceptive
  Locomotion Through Nonlinear Model-Predictive Control},'' \emph{IEEE
  Transactions on Robotics}, 8 2023. [Online]. Available:
  \url{http://arxiv.org/abs/2208.08373}
\BIBentrySTDinterwordspacing

\bibitem[Fankhauser(2018)]{Fankhauser2018}
P.~Fankhauser, ``{Perceptive Locomotion for Legged Robots in Rough Terrain},''
  \emph{ETH Zurich}, p. 177, 2018.

\bibitem[Agrawal et~al.(2022)Agrawal, Chen, Rai, and Sreenath]{Agrawal2021}
\BIBentryALTinterwordspacing
A.~Agrawal, S.~Chen, A.~Rai, and K.~Sreenath, ``{Vision-Aided Dynamic
  Quadrupedal Locomotion on Discrete Terrain Using Motion Libraries},''
  \emph{Proceedings - IEEE International Conference on Robotics and
  Automation}, pp. 4708--4714, 2022. [Online]. Available:
  \url{http://arxiv.org/abs/2110.00891}
\BIBentrySTDinterwordspacing

\bibitem[Deits and Tedrake(2015)]{Deits2015}
R.~Deits and R.~Tedrake, ``{Footstep planning on uneven terrain with
  mixed-integer convex optimization},'' \emph{IEEE-RAS International Conference
  on Humanoid Robots}, vol. 2015-Febru, pp. 279--286, 2015.

\bibitem[Griffin et~al.(2019)Griffin, Wiedebach, McCrory, Bertrand, Lee, and
  Pratt]{Griffin2019FootstepTerrain}
\BIBentryALTinterwordspacing
R.~J. Griffin, G.~Wiedebach, S.~McCrory, S.~Bertrand, I.~Lee, and J.~Pratt,
  ``{Footstep Planning for Autonomous Walking over Rough Terrain},''
  \emph{IEEE-RAS International Conference on Humanoid Robots}, vol. 2019-Octob,
  pp. 9--16, 7 2019. [Online]. Available: \url{http://arxiv.org/abs/1907.08673}
\BIBentrySTDinterwordspacing

\bibitem[Mastalli et~al.(2020)Mastalli, Havoutis, Focchi, Caldwell, and
  Semini]{Mastalli2020}
C.~Mastalli, I.~Havoutis, M.~Focchi, D.~G. Caldwell, and C.~Semini, ``{Motion
  planning for quadrupedal locomotion: Coupled planning, terrain mapping, and
  whole-body control},'' \emph{IEEE Transactions on Robotics}, vol.~36, no.~6,
  pp. 1635--1648, 2020.

\bibitem[Jenelten et~al.(2020)Jenelten, Miki, Vijayan, Bjelonic, and
  Hutter]{Jenelten2020}
F.~Jenelten, T.~Miki, A.~E. Vijayan, M.~Bjelonic, and M.~Hutter, ``{Perceptive
  Locomotion in Rough Terrain - Online Foothold Optimization},'' \emph{IEEE
  Robotics and Automation Letters}, vol.~5, no.~4, pp. 5370--5376, 2020.

\bibitem[Gangapurwala et~al.(2022)Gangapurwala, Geisert, Orsolino, Fallon, and
  Havoutis]{Gangapurwala2020}
\BIBentryALTinterwordspacing
S.~Gangapurwala, M.~Geisert, R.~Orsolino, M.~Fallon, and I.~Havoutis, ``{RLOC:
  Terrain-Aware Legged Locomotion Using Reinforcement Learning and Optimal
  Control},'' \emph{IEEE Transactions on Robotics}, vol.~38, no.~5, pp.
  2908--2927, 2022. [Online]. Available: \url{http://arxiv.org/abs/2012.03094}
\BIBentrySTDinterwordspacing

\bibitem[Siekmann et~al.(2021{\natexlab{a}})Siekmann, Godse, Fern, and
  Hurst]{Siekmann2020b}
\BIBentryALTinterwordspacing
J.~Siekmann, Y.~Godse, A.~Fern, and J.~Hurst, ``{Sim-to-Real Learning of All
  Common Bipedal Gaits via Periodic Reward Composition},'' 2021, pp.
  7309--7315. [Online]. Available: \url{http://arxiv.org/abs/2011.01387}
\BIBentrySTDinterwordspacing

\bibitem[Yu et~al.(2021)Yu, Jain, Escontrela, and Iscen]{Yu2021}
W.~Yu, D.~Jain, A.~Escontrela, and A.~Iscen, ``{Visual-Locomotion : Learning to
  Walk on Complex Terrains with Vision},'' \emph{5th Conference on Robot
  Learning (CoRL 2021)}, no. CoRL 2021, pp. 1--12, 2021.

\bibitem[Miki et~al.(2022)Miki, Lee, Hwangbo, Wellhausen, Koltun, and
  Hutter]{Miki2022LearningWild}
T.~Miki, J.~Lee, J.~Hwangbo, L.~Wellhausen, V.~Koltun, and M.~Hutter,
  ``{Learning robust perceptive locomotion for quadrupedal robots in the
  wild},'' \emph{Science Robotics}, vol.~7, no.~62, p. eabk2822, 2022.

\bibitem[Margolis et~al.(2021)Margolis, Chen, Paigwar, Fu, Kim, Kim, and
  Agrawal]{Margolis2021}
\BIBentryALTinterwordspacing
G.~B. Margolis, T.~Chen, K.~Paigwar, X.~Fu, D.~Kim, S.~Kim, and P.~Agrawal,
  ``{Learning to Jump from Pixels},'' no. CoRL 2021, pp. 1--10, 2021. [Online].
  Available: \url{http://arxiv.org/abs/2110.15344}
\BIBentrySTDinterwordspacing

\bibitem[Agarwal et~al.(2022)Agarwal, Kumar, Malik, and
  Pathak]{Agarwal2022LeggedVision}
\BIBentryALTinterwordspacing
A.~Agarwal, A.~Kumar, J.~Malik, and D.~Pathak, ``{Legged Locomotion in
  Challenging Terrains using Egocentric Vision},'' 11 2022. [Online].
  Available: \url{http://arxiv.org/abs/2211.07638}
\BIBentrySTDinterwordspacing

\bibitem[Rudin et~al.(2022{\natexlab{a}})Rudin, Hoeller, Bjelonic, and
  Hutter]{Rudin2022AdvancedEnd-to-End}
\BIBentryALTinterwordspacing
N.~Rudin, D.~Hoeller, M.~Bjelonic, and M.~Hutter, ``{Advanced Skills by
  Learning Locomotion and Local Navigation End-to-End},'' \emph{IEEE
  International Conference on Intelligent Robots and Systems}, vol. 2022-Octob,
  pp. 2497--2503, 9 2022. [Online]. Available:
  \url{http://arxiv.org/abs/2209.12827}
\BIBentrySTDinterwordspacing

\bibitem[Hoeller et~al.(2023)Hoeller, Rudin, Sako, and
  Hutter]{Hoeller2023ANYmalRobots}
\BIBentryALTinterwordspacing
D.~Hoeller, N.~Rudin, D.~Sako, and M.~Hutter, ``{ANYmal Parkour: Learning Agile
  Navigation for Quadrupedal Robots},'' 6 2023. [Online]. Available:
  \url{http://arxiv.org/abs/2306.14874}
\BIBentrySTDinterwordspacing

\bibitem[Loquercio et~al.(2022)Loquercio, Kumar, and
  Malik]{Loquercio2022LearningSupervision}
\BIBentryALTinterwordspacing
A.~Loquercio, A.~Kumar, and J.~Malik, ``{Learning Visual Locomotion with
  Cross-Modal Supervision},'' Tech. Rep., 2022. [Online]. Available:
  \url{http://arxiv.org/abs/2211.03785}
\BIBentrySTDinterwordspacing

\bibitem[Siekmann et~al.(2021{\natexlab{b}})Siekmann, Green, Warila, Fern, and
  Hurst]{Siekmann2021}
J.~Siekmann, K.~Green, J.~Warila, A.~Fern, and J.~Hurst, ``{Blind Bipedal Stair
  Traversal via Sim-to-Real Reinforcement Learning},'' \emph{Robotics: Science
  and Systems}, 2021.

\bibitem[Duan et~al.(2022)Duan, Malik, Gadde, Dao, Fern, and
  Hurst]{Duan2022LearningStones}
\BIBentryALTinterwordspacing
H.~Duan, A.~Malik, M.~S. Gadde, J.~Dao, A.~Fern, and J.~Hurst, ``{Learning
  Dynamic Bipedal Walking Across Stepping Stones},'' \emph{IEEE International
  Conference on Intelligent Robots and Systems}, vol. 2022-Octob, pp.
  6746--6752, 5 2022. [Online]. Available:
  \url{http://arxiv.org/abs/2205.01807}
\BIBentrySTDinterwordspacing

\bibitem[Castillo et~al.(2021)Castillo, Weng, Zhang, and Hereid]{Castillo2021}
\BIBentryALTinterwordspacing
G.~A. Castillo, B.~Weng, W.~Zhang, and A.~Hereid, ``{Robust Feedback Motion
  Policy Design Using Reinforcement Learning on a 3D Digit Bipedal Robot},''
  \emph{IEEE International Conference on Intelligent Robots and Systems}, pp.
  5136--5143, 2021. [Online]. Available: \url{http://arxiv.org/abs/2103.15309}
\BIBentrySTDinterwordspacing

\bibitem[Li et~al.(2023)Li, Peng, Abbeel, Levine, Berseth, and
  Sreenath]{Li2023RobustLearning}
\BIBentryALTinterwordspacing
Z.~Li, X.~B. Peng, P.~Abbeel, S.~Levine, G.~Berseth, and K.~Sreenath, ``{Robust
  and Versatile Bipedal Jumping Control through Multi-Task Reinforcement
  Learning},'' 2 2023. [Online]. Available:
  \url{http://arxiv.org/abs/2302.09450}
\BIBentrySTDinterwordspacing

\bibitem[Matthis and Fajen(2013)]{Matthis2013}
J.~S. Matthis and B.~R. Fajen, ``{Humans exploit the biomechanics of bipedal
  gait during visually guided walking over complex terrain},''
  \emph{Proceedings of the Royal Society B: Biological Sciences}, vol. 280, no.
  1762, 2013.

\bibitem[GIBSON(1958)]{GIBSON1958VisuallyAnimals}
\BIBentryALTinterwordspacing
J.~J. GIBSON, ``{Visually Controlled Locomotion and Visual Orientation in
  Animals},'' \emph{British Journal of Psychology}, vol.~49, no.~3, pp.
  182--194, 8 1958. [Online]. Available:
  \url{https://onlinelibrary.wiley.com/doi/10.1111/j.2044-8295.1958.tb00656.x}
\BIBentrySTDinterwordspacing

\bibitem[van Marum et~al.(2023)van Marum, Sabatelli, and
  Kasaei]{vanMarum2023LearningTerrain}
\BIBentryALTinterwordspacing
B.~van Marum, M.~Sabatelli, and H.~Kasaei, ``{Learning Perceptive Bipedal
  Locomotion over Irregular Terrain},'' Tech. Rep., 2023. [Online]. Available:
  \url{http://arxiv.org/abs/2304.07236}
\BIBentrySTDinterwordspacing

\bibitem[Rudin et~al.(2022{\natexlab{b}})Rudin, Hoeller, Reist, and
  Hutter]{Rudin2022LearningLearning}
\BIBentryALTinterwordspacing
N.~Rudin, D.~Hoeller, P.~Reist, and M.~Hutter, ``{Learning to Walk in Minutes
  Using Massively Parallel Deep Reinforcement Learning},'' in \emph{Proceedings
  of the 5th Conference on Robot Learning}, ser. Proceedings of Machine
  Learning Research, A.~Faust, D.~Hsu, and G.~Neumann, Eds., vol. 164.\hskip
  1em plus 0.5em minus 0.4em\relax PMLR, 3 2022, pp. 91--100. [Online].
  Available: \url{https://proceedings.mlr.press/v164/rudin22a.html}
\BIBentrySTDinterwordspacing

\bibitem[Fu et~al.(2022)Fu, Kumar, Agarwal, Qi, Malik, and Pathak]{Fu2021}
\BIBentryALTinterwordspacing
Z.~Fu, A.~Kumar, A.~Agarwal, H.~Qi, J.~Malik, and D.~Pathak, ``{Coupling Vision
  and Proprioception for Navigation of Legged Robots},'' \emph{Proceedings of
  the IEEE Computer Society Conference on Computer Vision and Pattern
  Recognition}, vol. 2022-June, pp. 17\,252--17\,262, 2022. [Online].
  Available: \url{http://arxiv.org/abs/2112.02094}
\BIBentrySTDinterwordspacing

\bibitem[Zhuang et~al.(2023)Zhuang, Fu, Wang, Atkeson, Schwertfeger, Finn, and
  Zhao]{Zhuang2023RobotLearning}
\BIBentryALTinterwordspacing
Z.~Zhuang, Z.~Fu, J.~Wang, C.~Atkeson, S.~Schwertfeger, C.~Finn, and H.~Zhao,
  ``{Robot Parkour Learning},'' Tech. Rep., 2023. [Online]. Available:
  \url{http://arxiv.org/abs/2309.05665}
\BIBentrySTDinterwordspacing

\bibitem[Haarnoja et~al.(2023)Haarnoja, Moran, Lever, Huang, Tirumala,
  Wulfmeier, Humplik, Tunyasuvunakool, Siegel, Hafner, Bloesch, Hartikainen,
  Byravan, Hasenclever, Tassa, Sadeghi, Batchelor, Casarini, Saliceti, Game,
  Sreendra, Patel, Gwira, Huber, Hurley, Nori, Hadsell, and
  Heess]{Haarnoja2023LearningLearning}
\BIBentryALTinterwordspacing
T.~Haarnoja, B.~Moran, G.~Lever, S.~H. Huang, D.~Tirumala, M.~Wulfmeier,
  J.~Humplik, S.~Tunyasuvunakool, N.~Y. Siegel, R.~Hafner, M.~Bloesch,
  K.~Hartikainen, A.~Byravan, L.~Hasenclever, Y.~Tassa, F.~Sadeghi,
  N.~Batchelor, F.~Casarini, S.~Saliceti, C.~Game, N.~Sreendra, K.~Patel,
  M.~Gwira, A.~Huber, N.~Hurley, F.~Nori, R.~Hadsell, and N.~Heess, ``{Learning
  Agile Soccer Skills for a Bipedal Robot with Deep Reinforcement Learning},''
  4 2023. [Online]. Available: \url{http://arxiv.org/abs/2304.13653}
\BIBentrySTDinterwordspacing

\bibitem[Fankhauser and Hutter(2016)]{Fankhauser2016a}
P.~Fankhauser and M.~Hutter, ``{A universal grid map library: Implementation
  and use case for rough terrain navigation},'' \emph{Studies in Computational
  Intelligence}, vol. 625, no. October 2017, pp. 99--120, 2016.

\bibitem[Fankhauser et~al.(2018)Fankhauser, Bloesch, and
  Hutter]{Fankhauser2018ProbabilisticLocalization}
\BIBentryALTinterwordspacing
P.~Fankhauser, M.~Bloesch, and M.~Hutter, ``{Probabilistic Terrain Mapping for
  Mobile Robots With Uncertain Localization},'' \emph{IEEE Robotics and
  Automation Letters}, vol.~3, no.~4, pp. 3019--3026, 10 2018. [Online].
  Available: \url{https://ieeexplore.ieee.org/document/8392399/}
\BIBentrySTDinterwordspacing

\bibitem[Plagemann et~al.(2009)Plagemann, Mischke, Prentice, Kersting, Roy, and
  Burgard]{Plagemann2009ALocomotion}
C.~Plagemann, S.~Mischke, S.~Prentice, K.~Kersting, N.~Roy, and W.~Burgard,
  ``{A bayesian regression approach to terrain mapping and an application to
  legged robor locomotion},'' \emph{Journal of Field Robotics}, vol.~26,
  no.~10, pp. 789--811, 10 2009.

\bibitem[Hoeller et~al.(2022)Hoeller, Rudin, Choy, Anandkumar, and
  Hutter]{Hoeller2022NeuralTerrain}
\BIBentryALTinterwordspacing
D.~Hoeller, N.~Rudin, C.~Choy, A.~Anandkumar, and M.~Hutter, ``{Neural Scene
  Representation for Locomotion on Structured Terrain},'' \emph{IEEE Robotics
  and Automation Letters}, vol.~7, no.~4, pp. 8667--8674, 6 2022. [Online].
  Available: \url{http://arxiv.org/abs/2206.08077}
\BIBentrySTDinterwordspacing

\bibitem[Yang et~al.(2023)Yang, Zhang, Geng, Wang, and
  Liu]{Yang2023Real-TimeEstimations}
B.~Yang, Q.~Zhang, R.~Geng, L.~Wang, and M.~Liu, ``{Real-Time Neural Dense
  Elevation Mapping for Urban Terrain with Uncertainty Estimations},''
  \emph{IEEE Robotics and Automation Letters}, vol.~8, no.~2, pp. 696--703, 2
  2023.

\bibitem[Schulman et~al.(2017)Schulman, Wolski, Dhariwal, Radford, and
  Klimov]{Schulman2017}
\BIBentryALTinterwordspacing
J.~Schulman, F.~Wolski, P.~Dhariwal, A.~Radford, and O.~Klimov, ``{Proximal
  Policy Optimization Algorithms},'' pp. 1--12, 2017. [Online]. Available:
  \url{http://arxiv.org/abs/1707.06347}
\BIBentrySTDinterwordspacing

\bibitem[Todorov et~al.(2012)Todorov, Erez, and Tassa]{Todorov2012}
E.~Todorov, T.~Erez, and Y.~Tassa, ``{MuJoCo: A physics engine for model-based
  control},'' \emph{IEEE International Conference on Intelligent Robots and
  Systems}, pp. 5026--5033, 2012.

\bibitem[Ma et~al.(2023)Ma, Farshidian, and Hutter]{Ma2023LearningManipulators}
\BIBentryALTinterwordspacing
Y.~Ma, F.~Farshidian, and M.~Hutter, ``{Learning Arm-Assisted Fall Damage
  Reduction and Recovery for Legged Mobile Manipulators},'' \emph{IEEE
  International Conference on Robotics and Automation, 2023}, 3 2023. [Online].
  Available: \url{http://arxiv.org/abs/2303.05486}
\BIBentrySTDinterwordspacing

\bibitem[Ronneberger et~al.(2015)Ronneberger, Fischer, and
  Brox]{Ronneberger2015U-net:Segmentation}
O.~Ronneberger, P.~Fischer, and T.~Brox, ``{U-net: Convolutional networks for
  biomedical image segmentation},'' \emph{MICCAI 2015}, vol. 9351, pp.
  234--241, 2015.

\end{thebibliography}

\end{document}